\documentclass[letterpaper, 10 pt, conference]{ieeeconf}

\IEEEoverridecommandlockouts                              

\usepackage{graphicx} 
\usepackage{amsmath} 
\usepackage{multirow}
\usepackage{floatrow}
\usepackage{cite}
\usepackage[table]{xcolor}
\usepackage{multicol}
\newfloatcommand{capbtabbox}{table}[][\FBwidth]
\usepackage{bm}
\usepackage{siunitx}
\usepackage{amsfonts}
\let\labelindent\relax
\usepackage[inline]{enumitem}
\usepackage{soul}
\usepackage{breqn}
\usepackage[switch]{lineno}
\usepackage{svg}

\DeclareMathOperator*{\argmin}{arg\,min}
\newcommand{\ubar}[1]{\text{\b{$#1$}}}

\title{Deep Dynamics: Vehicle Dynamics Modeling with a Physics-Constrained Neural Network for Autonomous Racing
}

\author{John Chrosniak$^{1}$, Jingyun Ning$^{1}$, and Madhur Behl$^{1}$%
\thanks{$^{1}$Department of Computer Science, University of Virginia, Charlottesville, USA
        {\tt\footnotesize \{jlc9wr, jn2ne, madhur.behl\}@virginia.edu}}%
~}

\begin{document}

\maketitle

\begin{abstract}


Autonomous racing is a critical research area for autonomous driving, presenting significant challenges in vehicle dynamics modeling, such as balancing model precision and computational efficiency at high speeds ($>$280 km/h), where minor errors in modeling have severe consequences.
Existing physics-based models for vehicle dynamics require elaborate testing setups and tuning, which are hard to implement, time-intensive, and cost-prohibitive.
Conversely, purely data-driven approaches do not generalize well and cannot adequately ensure physical constraints on predictions.
This paper introduces Deep Dynamics, a physics-constrained neural network (PCNN) for autonomous racecar vehicle dynamics modeling. 
It merges physics coefficient estimation and dynamical equations to accurately predict vehicle states at high speeds.
A unique Physics Guard layer ensures internal coefficient estimates remain within their nominal physical ranges.
Open-loop and closed-loop performance assessments, using a physics-based simulator and full-scale autonomous Indy racecar data, highlight Deep Dynamics as a promising approach for modeling racecar vehicle dynamics.

\end{abstract}



\section{Introduction}


As the field of autonomous vehicles has grown, the majority of research has been directed towards common driving situations. 
However, to enhance safety and performance, it is crucial to address the challenges presented by extreme driving environments. 
A quickly progressing area of research, autonomous racing, offers a valuable setting for this exploration. 
Recent advances in autonomous racing have been summarized comprehensively in~\cite{betz2022autonomous}.
Inspired by high-speed racing, competitions such as the Indy Autonomous Challenge~\cite{ wischnewski2022indy}, F1tenth~\cite{o2019f1}, and Formula Driverless~\cite{kabzan2019learning} have emerged where scaled and full-size driverless racecars compete to achieve the fastest lap times and overtake each other at speeds exceeding $280$ km/h.
The control software for autonomous racing must navigate racecars at extreme speeds and dynamic limits, a feat easily handled by skilled human racers but challenging for current autonomous driving systems.

An essential aspect of autonomous racing is the precise modeling of a vehicle's dynamic behavior. 
This encompasses using linearized or nonlinear differential equations to capture motion, tire dynamics, aerodynamics, suspension, steering, and drivetrain physics. 
While kinematic models focus on vehicle geometry, dynamic models delve deeper, estimating the acting forces to predict future movement, especially crucial in racing's nonlinear regions characterized by rapid vehicle state changes and elevated dynamic loads.
Obtaining precise coefficient values for modeling components like tires, suspension systems, and drivetrains is crucial, but extremely challenging. 
Determining tire coefficients (e.g. Pacejka coefficients\cite{pacejka}) involves extensive testing with specialized equipment (tire rigs) and significant time \cite{flattrac}. 
Additionally, calculating a racecar's moment of inertia is a laborious task necessitating vehicle disassembly, precise component weighing and positioning, and the use of CAD software or multi-body physics simulations. 
Frequent recalculations are also necessary to accommodate varying vehicle setups, track conditions, and tire wear.

Deep neural networks (DNNs) have proven adept at capturing complex nonlinear dynamical effects~\cite{weiss2020deepracing}, offering a simpler alternative to physics-based models as they obviate the need for extra testing equipment. 
Nevertheless, their substantial computational demands limit their integration for real-time model-predictive control (MPC), and they can often generate outputs unattainable by the actual system.
Consequentially, this has sparked increasing interest in methods that combine DNNs with physics-based approaches. 
For autonomous vehicles, an example of this approach is the Deep Pajecka Model~\cite{kim2022physics}.
However, this model has limitations such as reliance on sampling-based control, unconstrained Pacejka coefficient estimations, and not being tested on real data. 
We present Deep Dynamics, a physics-constrained neural network (PCNN) that to our best knowledge is the first model of its kind suited for vehicle dynamics modeling in autonomous racing - a domain characterized by high accelerations and high speeds.
This paper makes the following research contributions: 

\begin{enumerate}[noitemsep,topsep=0pt]
    \item We introduce Deep Dynamics - A PCNN that can estimate the Pajecka tire coefficients, drivetrain coefficients, and moment of inertia that capture the complex and varying dynamics of an autonomous racecar.
    \item We introduce the Physics Guard layer designed to constrain estimated coefficients to lie within their physically meaningful ranges and enforce the physics governing vehicle dynamics, hence the name ``physics-constrained''.
    \item We examine both open-loop and closed-loop performance, utilizing a mix of simulation and real-world data from a full-scale autonomous racing car competing in the Indy Autonomous Challenge.
\end{enumerate}

 Our findings affirm that Deep Dynamics can effectively integrate with model-predictive control-based (MPC) trajectory following methods, all while leveraging the advantages of data-driven vehicle modeling without the complications associated with physics-based modeling.

\section{Related Work}
\label{ssec:rw}

Vehicle dynamics modeling is a thoroughly investigated subject for both passenger autonomous driving and autonomous racing. 
Physics-based vehicle dynamics models can vary greatly in complexity, ranging from simple point-mass to single-track (bicycle) models, to complex multi-body models \cite{commonroad}. 
Kinematic and dynamic single-track models are commonly used for vehicle dynamics modeling because they offer a good balance of simplicity and accuracy~\cite{7039455}. 
However, the accuracy of such models correlates with the accuracy of the model coefficients, which are often difficult to identify and tune \cite{9098865}.
Researchers have explored a variety of system identification approaches to learn the parameters of physics models through observations of the vehicle's motion. 
\cite{6882826} describes a method to estimate the coefficients for a vehicle's drivetrain model and \cite{pacejkasysid} focuses on identifying tire coefficients. 
Such methods often make a simplifying assumption that the model coefficients are time-invariant, which is not valid in autonomous racing as coefficients may vary due to changes in racecar setups or tire wear. 
Hybrid models that use a combination of data-driven and physics-based approaches have also been proposed. 
\cite{JainBayesRace2020, 8754713, ning2023scalable} use Gaussian Processes to correct the predictions of a simple physics model to match observed states.
\cite{COSTA2023104469} addresses the same problem using a DNN. 
These approaches capture time-variant dynamics but lack interpretability and do not ensure that corrected predictions will make physical sense.




Purely data-driven models, like in \cite{scirobotics.aaw1975} and \cite{7989202}, learn vehicle dynamics using historical observations of the vehicle's state and control inputs to train a DNN. 
\cite{9357196} proposed the use of a recurrent neural network with a Gated Recurrent Unit (GRU) mechanism to accomplish the same task. 
These methods have shown the ability to perform well within the distribution of data they were trained on but, as is the case with supervised machine learning approaches, they struggle to generalize on out-of-distribution data. The complexity and non-linearity of DNNs also limit the use of such models for model-based predictive control. 
Additionally, while these models are great at learning complex patterns in data, they are not designed to obey the laws of physics and can produce state predictions that the vehicle cannot physically realize.


Recently, methods combining dynamical equations with deep learning for vehicle dynamics have been gaining traction as they adeptly merge the strengths of both approaches. 
In the study by \cite{baier2022hybrid}, hybrid physics and deep learning models were employed to model lateral dynamics in ships and quadrotors. Similarly, \cite{jmse10020148} trained a physics-informed neural network (PINN) to determine the dynamics of an unmanned surface vehicle. 
The closest related work to ours is the Deep Pacejka Model (DPM) introduced in \cite{kim2022physics}, which estimates tire coefficients for autonomous vehicles based on historical states and control inputs. 
However, as detailed in Section \ref{ssec:dpm}, the DPM bears limitations that make it unfit for autonomous racing. 
Our research aims to address these limitations, introducing a new PCNN model suitable for a real autonomous racecar.

\section{Problem Statement}

\begin{table}
    \vspace{6pt}
    \centering
    \caption{Dynamic Single-Track Vehicle Model}
    \resizebox{\columnwidth}{!}{
    \begin{tabular}{l l l}
        \hline
        \textbf{State Variable} & \textbf{Notation} & \textbf{State Equation}  \\
         Horizontal Position (\si{\meter}) & $x$ & $x_{t+1}$ = $x_t + (v_{x_t}\cos\theta_t - v_{y_t}\sin\theta_t)T_s$ \\
         Vertical Position (\si{\meter}) & $y$ & $y_{t+1}$ = $y_t + (v_{x_t}\sin\theta_t + v_{y_t}\cos\theta_t)T_s$ \\
         Inertial Heading (\si{\radian}) & $\theta$ & $\theta_{t+1} = \theta_t + (\omega_t)T_s$ \\
         Longitudinal Velocity (\si{\meter\per\second}) & $v_x$ & $v_{x_{t+1}} = v_{x_t} + \frac{1}{m}(F_{rx}-F_{fy}\sin\delta_t + mv_{y_t}\omega_t)T_s$ \\[3pt]
         Lateral Velocity (\si{\meter\per\second}) & $v_y$ & $v_{y_{t+1}} = v_{y_t} + \frac{1}{m}(F_{ry}+F_{fy}\cos\delta_t - mv_{x_t}\omega_t)T_s$ \\[3pt]
         Yaw Rate (\si{\radian\per\second}) & $\omega$ & $\omega_{t+1} = \omega_t + \frac{1}{I_z}(F_{fy}l_f\cos\delta_t - F_{ry}l_r)T_s$ \\
         Throttle (\%) & $T$ & $T_{t+1} = T_t + \Delta T$ \\
         Steering Angle (\si{\radian}) & $\delta$ & $\delta_{t+1} = \delta_t + \Delta\delta$ \\
         \hline
         \textbf{Input Variable} & \multicolumn{2}{l}{\textbf{Notation}} \\
         Throttle Change (\%) & $\Delta T$ \\
         Steering Change (\si{\radian}) & $\Delta\delta$ \hspace{2pt} \\
         \hline 
         \multicolumn{3}{c}{\textbf{Drivetrain Model}} \hspace{2pt}\\ 
         \multicolumn{3}{c}{$F_{rx} = (C_{m1} - C_{m2}v_x)T - C_{r0} - C_d{v_x}^2$} \hspace{2pt} \\
         \hline
         \multicolumn{3}{c}{\textbf{Pacejka Tire Model}} \hspace{2pt} \\ 
         \multicolumn{3}{c}{$\alpha_f = \delta - \arctan{\frac{\omega l_f + v_y} {v_x}} + G_{f}$ \quad $\alpha_r = \arctan{\frac{\omega l_r - v_y}{v_x}} + G_{r}$} \\
         \multicolumn{3}{c}{$F_{fy} = K_{f} + D_f\left(\sin\left(C_f\arctan\left(B_f\alpha_f - E_f\left(B_f\alpha_f - \arctan\left(B_f\alpha_f\right)\right)\right)\right)\right)$}  \\ 
         \multicolumn{3}{c}{$F_{ry} = K_{r} + D_r\left(\sin\left(C_r\arctan\left(B_r\alpha_r - E_r\left(B_r\alpha_r - \arctan\left(B_r\alpha_r\right)\right)\right)\right)\right)$}  \hspace{2pt} \\[2pt]
         \hline
    \end{tabular}}
    \label{tab:dynamics_equations}
\end{table}

We begin with a brief overview of vehicle dynamics followed by the problem formulation.

\subsection{Dynamic Single-Track Vehicle Model}
\label{ssec:vehicle_dynamics}

\begin{figure}
\vspace{4pt}
    \centering
    \includegraphics[width=\columnwidth]{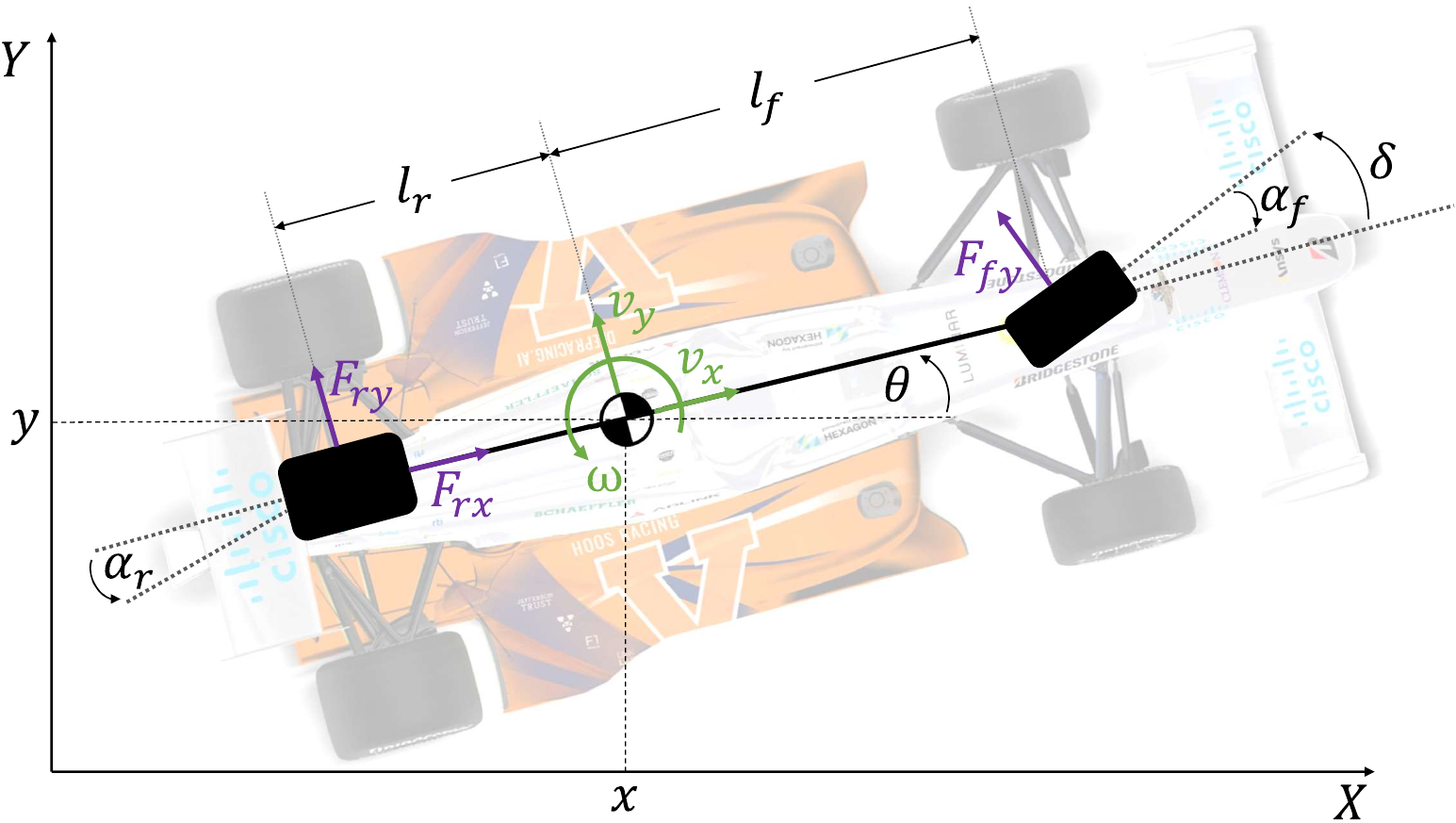}
    \caption{The dynamic single-track vehicle model.}
    \label{fig:bicycle-model}
\end{figure}

We model the autonomous racecar using a dynamic single-track vehicle model~\cite{7039455}, with the state variables, discrete-time state equations, and notations listed in Table~\ref{tab:dynamics_equations}.
The accompanying free-body diagram for this model is shown in Figure \ref{fig:bicycle-model}. 
At time $t$, the states of the system are $S_t = [x_t, y_t, \theta_t, v_{x_t}, v_{y_t}, \omega_t, T_t, \delta_t] \in \mathbb{R}^8$, and the control inputs $U_t$ consist of changes to the vehicle's throttle and steering i.e. $U_t = [\Delta T_t, \Delta\delta_t] \in \mathbb{R}^2$. 
The single-track model (also known as the bicycle model) makes the assumption that the dynamics of the car can be described
based on two virtual wheels located along the longitudinal axis of the racecar.
A drivetrain model \cite{JainBayesRace2020} estimates the longitudinal force at the rear wheels, $F_{rx}$, using drivetrain coefficients, $C_{m1}$ and $C_{m2}$, the rolling resistance $C_{r0}$, and the drag resistance $C_d$.
$C_{m1}$ (N) linearly links throttle, $T$, to rear-wheel force and $C_{m2}$ (kg/s) models drivetrain damping.
Drag and rolling resistance forces are simplified as $F_{d}=C_dv_x^2$ and $F_{ro}=C_{r0}$ respectively, with coefficients $C_d$ ($\si{\kilo\gram\per\meter}$) and $C_{r0}$ ($\si{\newton}$).


The Pacejka Magic Formula tire model \cite{pacejka} estimates the lateral tire forces on the front and rear wheels, $F_{fy}$ and $F_{ry}$,
inflicted by the vehicle’s movement and steering. 
The Pacejka model first calculates the sideslip angles for the front and rear wheels, $\alpha_f$ and $\alpha_r$, using state $S_t$ and the distance between the vehicle's center of gravity to the front and rear axles, $l_f$ and $l_r$. 
The sideslip angles and sets of tire coefficients for the front and rear wheels, $(B, C, D, E, G, K)_{f/r}$, are then used to estimate the lateral tire forces (Table~\ref{tab:dynamics_equations}). 
Additional model coefficients include the mass, $m$, and moment of inertia, $I_z$, of the vehicle.
The set of model coefficients $\Phi = \{ m, l_f, l_r, B_{f/r}, C_{f/r}, D_{f/r}, E_{f/r}, G_{f/r}, K_{f/r},\linebreak C_{m1}, C_{m2}, C_{r0}, C_d, I_z\} \in \mathbb{R}^{20}$ together with the state $S_t$ and the control input $U_t$ fully describe the system's dynamics.

\noindent \textbf{Modeling Assumptions.} We make the following modeling assumptions for the dynamic single-track model used in this work: 
\begin{enumerate*}[noitemsep,topsep=0pt]
    \item The model assumes that the racecar drives in 2D space despite the banking on real-world racetracks. 
    \item We assume that there is no significant actuation delay for the control inputs $\Delta T$, and $\Delta \delta$. 
    \item We assume that the vehicle's mass, $m$, and geometric properties, $l_r$ and $l_f$, belong to the set of known coefficients, $\Phi_k$, because they can be measured in a typical garage. The remaining coefficients belong to the set of unknown coefficients, $\Phi_u$, because require cost and time-intensive procedures to obtain.
Therefore, $\Phi = \{ \Phi_{k} \cup \Phi_{u} \} $, where $\Phi_{k} = \{m, l_f, l_r\} $ and $\Phi_{u} = \{B_{f/r}$, $C_{f/r}$, $D_{f/r}$, $E_{f/r}$, $G_{f/r}$, $K_{f/r}$, $C_{m1}$, $C_{m2}$, $C_{r0}$, $C_d$, $I_z\}$. 
Although the precise values of $\Phi_u$ are unknown, we posit that each coefficient is bounded within a known nominal range, denoted as \textbf{$\ubar{\Phi}_u \leq \Phi_u \leq \bar{\Phi}_u$}.
\end{enumerate*}


\subsection{Vehicle Dynamics Identification Problem}

In line with the definitions in~\cite{kim2022physics, 9357196}, the evolution of the model states $S_t = [x_t, y_t, \theta_t, v_{x_t}, v_{y_t}, \omega_t, T_t, \delta_t]$ is governed by the states $X_t = [v_{x_t}, v_{y_t}, \omega_t, T_t, \delta_t] \in \mathbb{R}^5$ and the control inputs $U_t = [\Delta T_t, \Delta\delta_t]$. 
Hence, we exclusively utilize the state vector $X_t$ moving forward.
Given a dataset $\mathcal{D} = [[X_1, U_1], [X_2, U_2], ..., [X_N, U_N]]$, with $N$ state-input observations, we first split the data into two disjoint sets $\mathcal{D} = \{\mathcal{D}^{\text{train}} \cup \mathcal{D}^{\text{test}}\}$, with $\mathcal{D}^{\text{train}}$ used for model identification and $\mathcal{D}^{\text{test}}$  for validation.
Our goal is to form a model $f$ that, given the current state $X_t$, input $U_t$, known coefficients $\Phi_k$, and estimated coefficients $\hat{\Phi}_{u_t}$, predicts the next state:
\begin{equation}
    \hat{X}_{t+1} = f(X_t, U_t, \Phi_{k}, \hat{\Phi}_{u_t})
\end{equation}
Since $f$ is realized through the state equations in Table \ref{tab:dynamics_equations}, the error between the model output $\hat{X}_{t+1}$ and the observed ${X}_{t+1}$ is governed by the unknown coefficient estimates $\hat{\Phi}_{u_t}$.
Given the nominal ranges for the unknown model coefficients $\ubar{\Phi}_u \leq \Phi_u \leq \bar{\Phi}_u$, model $f$, training data $\mathcal{D}^{\text{train}}$ with $N^{\text{train}}$ samples, the formal problem can be stated as:

\begin{equation}
    \argmin\limits_{\hat{\Phi}_{u_t}}\, \frac{1}{N^{\text{train}}}\sum_{i=1}^{N^{\text{train}}}(X_i - \hat{X}_i)^2\; \text{s.t.}\; \ubar{\Phi}_u \leq \hat{\Phi}_u \leq \bar{\Phi}_u
    \label{eq:problem}
\end{equation}

We approach the estimation of $\hat{\Phi}_{u_t}$ through a PCNN that is intrinsically informed by the nominal ranges of $\Phi_u$.


\section{Deep Dynamics}

Hybrid models combining physics-based and data-driven approaches have emerged as effective methods for modeling non-linear dynamic behaviors. The Deep Pacejka Model (DPM) \cite{kim2022physics} uses hybrid models for autonomous driving, yet its limitations are pronounced when applied to the context of autonomous racing.

\subsection{Limitations of the Deep Pacejka Model} \label{ssec:dpm}

The DPM uses a neural network to only estimate the Pacejka coefficients $\{B_{f/r}, C_{f/r}, D_{f/r}, E_{f/r}\} \subset \Phi_u$, and rather than using a physics-based drivetrain model, directly estimates the longitudinal force, $F_{rx}$, for an autonomous vehicle given a history of the vehicle's state and control inputs. 
The DNN outputs are then used with the dynamic bicycle model state equations to predict the vehicle's subsequent observed state, $X_{t+1}$. 
Backpropagating results in a network that learns Pacejka coefficients and a purely data-driven drivetrain model that capture the vehicle's dynamics.
\begin{figure}
\vspace{4pt}
    \centering
    \includegraphics[width=\columnwidth]{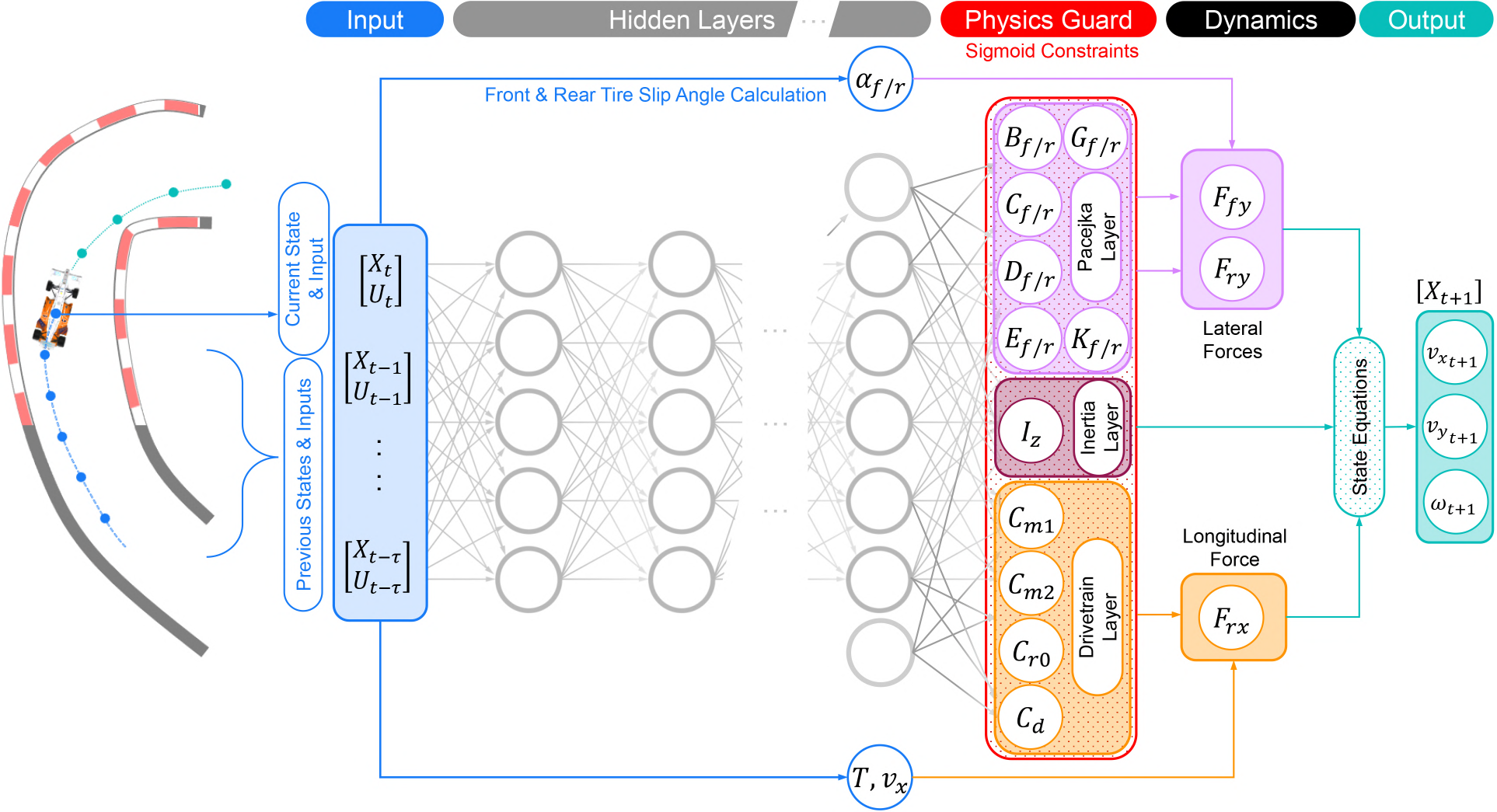}
    \caption{Deep Dynamics uses a DNN to learn a vehicle's single-track model coefficients from historical states and control inputs.}
    \label{fig:ddm}
\end{figure}

The DPM suffers from several limitations. First, the DPM solves an unconstrained version of the problem in Equation \ref{eq:problem}, i.e. there are no bounds on $\hat{\Phi}_u$. 
This can cause DPM to produce coefficients outside of the nominal ranges $\ubar{\Phi}_u$ and $\bar{\Phi}_u$, which could result in non-sensical predictions of the vehicle's state. Second, the DPM assumes precise knowledge of the vehicle's moment of inertia, $I_z$. This is a very strong assumption because the precise value of $I_z$ requires knowledge of the weight distribution of the car. In addition, the DPM does not have a drivetrain model as there is no learned relation between the vehicle throttle, $T$, and longitudinal wheel force, $F_{rx}$, which limits the DPM to using PID throttle control to track $V_{des}$ set by sampling-based MPC rather than directly controlling $T$. More importantly, the Deep Pacejka Model was designed for low-speed autonomous driving and not for a high-speed racing environment.

\subsection{Deep Dynamics}

We propose the Deep Dynamics Model (DDM), a PCNN capable of estimating the unknown coefficients for a single-track dynamical model (Section \ref{ssec:vehicle_dynamics}) that are guaranteed to lie within their physically meaningful ranges, $\ubar{\Phi}_u$ and $\bar{\Phi}_u$. The model architecture for the DDM is shown in Figure \ref{fig:ddm}. The DDM is given a history of length $\tau$ of vehicle states and control inputs, $[[X_{t-\tau}, U_{t-\tau}],...,[X_t,U_t]]$, which is then fed through the hidden layers of a DNN. 
We have introduced a Physics Guard layer which applies a Sigmoid activation function to the output, $z$, of the last hidden layer and scales the result to produce $\hat{\Phi}_{u_t}$ between the bounds $\ubar{\Phi}_u$ and $\ubar{\Phi}_u$ using

\begin{equation}
    \hat{\Phi}_u = \sigma(z) \cdot (\bar{\Phi}_u - \ubar{\Phi}_u) + \ubar{\Phi}_u
\end{equation}

A Sigmoid function was used for simplicity, but this mechanism could be implemented using any activation function with finite boundaries. The estimated coefficients and single-track model state equations are then used to predict $\hat{X}_{t+1}$, and the backpropagated loss function is computed using 

\begin{equation}
    \mathcal{L}({\hat{X}}) = \frac{1}{3}\sum_{i=1}^3(X^{(i)}_{t+1} - \hat{X}^{(i)}_{t+1})^2
\end{equation}

across the state variables $[v_x, v_y, \omega]$. The states $T$ and $\delta$ were not used in the loss function as the change in these states is determined by the control inputs $\Delta T$ and $\Delta \delta$.
In contrast to the DPM, the architecture of the DDM allows for longitudinal control using an MPC solver due to the inclusion of a physics-based drivetrain model. Additionally, the DDM does not assume prior knowledge of the vehicle's moment of inertia, $I_z$. Finally, the Physics Guard prevents the DDM from estimating non-sensical single-track model coefficients.

\subsection{Model-Predictive Control with Deep Dynamics}
\label{ssec:mpc}

For closed-loop testing, we use MPC to control the vehicle's steering and throttle.
An optimal raceline, $W \in \mathbb{R}^{n\times 2}$, containing a sequence of $n$ reference points $w_i = [x_{\text{ref},i}, y_{\text{ref},i}] \in \mathbb{R}^2$ is tracked using MPC by solving for the control inputs, $U_0,...,U_{H-1}$, that minimize the cost function

\begin{equation}
\label{eq:mpc}
\begin{split}
    \argmin\limits_{U_0,...,U_{H-1}}\, \sum_{h=1}^{H} \begin{Vmatrix}
        x_h - x_{\text{ref},h} \\
        y_h - y_{\text{ref},h}
    \end{Vmatrix}_Q + \sum_{h=0}^{H-1} \begin{Vmatrix}
        \Delta T_h \\
        \Delta \delta_h
    \end{Vmatrix}_R
\end{split}
\end{equation}

across the forward prediction horizon $H$. 
The first term, the tracking cost, penalizes deviations from the optimal raceline using the cost matrix $Q \in \mathbb{R}^{2\times 2}$. The second term, the actuation cost, penalizes rapid steering and throttle changes using the cost matrix $R \in \mathbb{R}^{2\times 2}$. Once the optimal series of control inputs $U_0,...,U_{H-1}$ have been derived, the controller can enact $U_0$ and repeat this procedure.

At time $t$, Deep Dynamics is used to predict the vehicle's trajectory over the forward horizon, $H$, by first estimating model coefficients $\hat{\Phi}_{u_t}$ using the observed states and control inputs from times $t-\tau$ to $t$. These coefficients are held constant and used by MPC to propagate the predicted state of the vehicle from time $t$ to $t+H$. Holding $\hat{\Phi}_{u_t}$ constant throughout $H$ allows an MPC solver to compute steering and throttle commands that minimize Equation \ref{eq:mpc} without having to differentiate through the DDM's hidden layers, making it feasible to run in real-time.



\section{Experiments \& Results}

\subsection{Training and Testing Datasets}

\begin{figure}
    \centering
    \vspace{4pt}
    \includegraphics[width=\columnwidth]{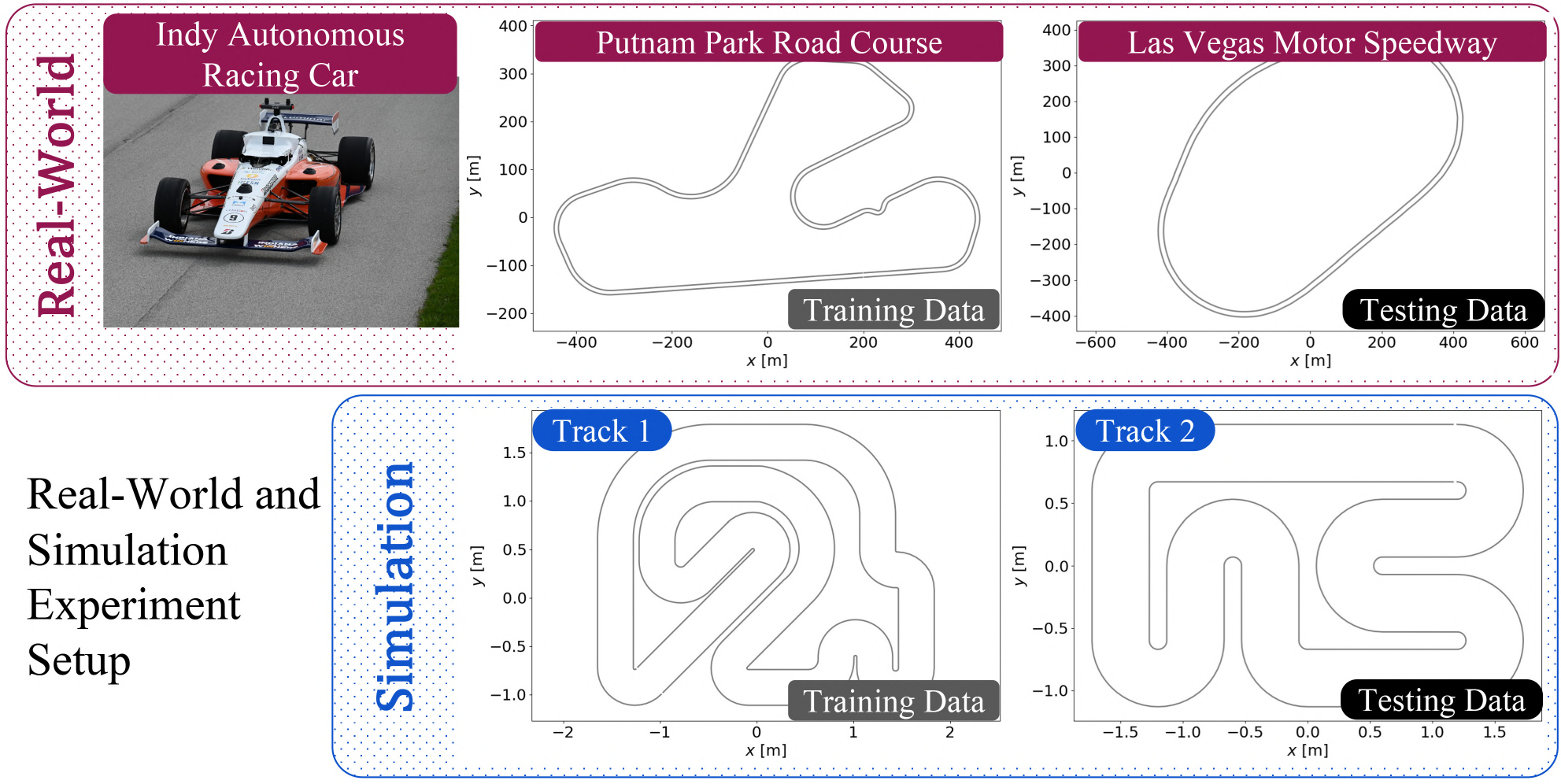}
    \caption{A real-world, full-scale autonomous racecar was used to collect data at the Putnam Park Road Course and Las Vegas Motor Speedway. In simulation, laps were driven on Track 1 and Track 2.}
    \label{fig:tracks}
\end{figure}

The open-loop performance of the DDM and DPM was evaluated on both real and simulated data (shown in Figure \ref{fig:tracks}). 
Real world data $\mathcal{D}_{\text{real}}$ was compiled from a full-scale Indy autonomous racecar \cite{racecar2023}. 
State information was sampled at a rate of 25 Hz from a 100 Hz Extended Kalman Filter that fuses measurements from two GNSS signals with RTK corrections and their IMU units. 
The training data, $\mathcal{D}^{\text{train}}_{\text{real}}$, consisting of 13,418 samples was collected at the Putnam Park road course in Indianapolis, while
the test set, $\mathcal{D}^{\text{test}}_{\text{real}}$, was curated from laps at the Las Vegas Motor Speedway and consists of 10,606 samples. 
Simulation dataset, $\mathcal{D}_{\text{sim}}$, contains ground-truth vehicle state information from laps driven in a 1:43 scale racecar simulator \cite{JainBayesRace2020}, where the ground-truth coefficient values $\Phi$ are known.
Data was collected from two different track configurations at 50 Hz using pure-pursuit to drive the vehicle along a raceline. 
Track 1 was used to form the training dataset, $\mathcal{D}^{\text{train}}_{\text{sim}}$, and Track 2 was used for the testing dataset, $\mathcal{D}^{\text{test}}_{\text{sim}}$, with both having 1,000 samples.

\subsection{Nominal Model Coefficient Ranges}

\begin{table}[]
    \vspace{6pt}
    \centering
    \resizebox{0.8\textwidth}{!}{
    \begin{tabular}{l l l l l}
        \hline
         & \multicolumn{2}{c}{\textbf{Sim Data}} & \multicolumn{2}{c}{\textbf{Real Data}}  \\
        \hline
        \textbf{Coefficient} & \textbf{Min} & \textbf{Max} & \textbf{Min} & \textbf{Max} \\
        \hline
        $B_{f/r}$  &  5.0 & 30.0 & 5.0 & 30.0 \\
        $C_{f/r}$  &  0.5 & 2.0 & 0.5 & 2.0 \\
        $D_{f/r}$  &  0.1 & 1.9 & 100.0 & 10000.0 \\
        $E_{f/r}$  &  -2.0 & 0.0 & -2.0 & 0.0 \\
        $K_{f/r}$  & -0.003 & 0.003 & -300.0 & 300.0  \\
        $G_{f/r}$  &  -0.02 & 0.02 & -0.02 & 0.02 \\
        $C_{m1}\:(\si{\newton})$  &  0.1435 & 0.574 & 500.0 & 2000.0 \\
        $C_{m2}\:(\si{\kilo\gram\per\second})$  &  0.0273 & 0.109 & 0.0 & 1.0 \\
        $C_{r0}\:(\si{\newton})$  &  0.0259 & 0.1036 & 0.1 & 1.4 \\
        $C_d\:(\si{\kilo\gram\per\meter})$  &  1.75e-4 & 7.0e-4 & 0.1 & 1.0 \\
        $I_z\:(\si{\kilo\gram\meter\squared})$  &  1.39e-5 & 5.56e-5 & 500.0 & 2000.0 \\
        \hline
    \end{tabular}}
    \caption{Coefficient Ranges for the Dynamic Single-Track Model}
    \label{tab:coeff_ranges}
\end{table}

The Physics Guard layer requires bounds $\ubar{\Phi}_u$ and $\bar{\Phi}_u$ for each model coefficient ${\Phi}_u$ estimated by Deep Dynamics.
Table {\ref{tab:coeff_ranges}} outlines the coefficient ranges for both simulation and real-world data.
The ranges for Pacejka coefficients were set using the bounds described in \cite{pacejkaparams}.
For the scaled-down simulation vehicle, $D_{f/r}$'s range was adjusted to reflect a $1$ $\si{\newton}$ normal force. 
The simulator's acceptable range dictated the adjustment for $K_{f/r}$.
Model coefficients in the simulator are known, so ranges for the simulation vehicle's drivetrain coefficients and moment of inertia were created artificially by halving and doubling the ground-truth values. 
For the real vehicle, the range for the moment of inertia, $I_z$, was set using an estimate of $1,000$ $\si{\kilo\gram\meter\squared}$ provided by Dallara.
The ranges for the drag resistance, $C_d$, and rolling resistance, $C_{r0}$, were determined by halving and doubling estimated values from \cite{beckman1998physics}.  
The range for drivetrain coefficient $C_{m1}$ was set using the engine specification for Indy cars and the coefficient range for $C_{m2}$ was set to account for minor damping.

\subsection{Evaluation Metrics}

The root-mean-square error (RMSE) and maximum error ($\epsilon_{max}$) for the predicted longitudinal velocity, $v_x$, lateral velocity, $v_y$, and yaw rate, $\omega$, were measured to compare each model's one step prediction performance. 
In addition, we also generate horizon predictions using the estimated coefficients of each model and compute the average displacement error (ADE) and final displacement error (FDE) \cite{7780479}. 
A time horizon of 300ms was used for generating trajectories on $\mathcal{D}^{\text{test}}_{\text{sim}}$ and a horizon of 600ms was used for $\mathcal{D}^{\text{test}}_{\text{real}}$. 


\subsection{Open-Loop Testing}

To highlight the limitation of presuming precise knowledge of the moment of inertia, $I_z$, we trained three distinct iterations of the DPM: one given the ground truth value, termed DPM (GT); another with a value inflated by 20\%, labeled DPM (+20\%); and a third with a value reduced by 20\%, referred to as DPM (-20\%).
The DPM (GT) offers a comparison against the best possible DPM while the others provide a comparison against more realistic models. 
While the choice of a $20\%$ deviation is arbitrary, it serves to emulate real-world scenarios where the exact value of $I_z$ is unknown.
\begin{table}
    \centering
    \resizebox{\textwidth}{!}{
    \begin{tabular}{c l c c c c c c c c}
        \hline
         && \multicolumn{2}{c}{\boldmath$v_x \:(m/s)$} & \multicolumn{2}{c}{\boldmath$v_y \: (m/s)$} & \multicolumn{2}{c}{\boldmath$\omega \: (rad/s)$} & \multicolumn{2}{c}{\textit{\textbf{Displacement (m)}}}\\
         \cline{3-10}
         && RMSE & $\epsilon_{max}$ & RMSE & $\epsilon_{max}$ & RMSE & $\epsilon_{max}$ & ADE & FDE \\
         \hline
         \multirow{4}{*}{\shortstack{\textbf{Sim} \\ \textbf{Data}}} & DPM (GT) & 0.0343 & 0.2017 & 0.1252 & 0.6762 & 4.306 & 30.606 & 0.0302  & 0.0942 \\
         & DPM (+20\%) & 0.0270 & 0.2030 & 0.0889 & 0.5625 & 2.747 & 23.930  & 0.0263 & 0.0764 \\
         & DPM (-20\%) & 0.0331 & 0.2012 & 0.1202 & 0.6814 & 3.578 & 35.892 & 0.0467 & 0.1280 \\
         & DDM (ours) & \textbf{1.506e-5} & \textbf{1.051e-4} & \textbf{1.839e-4} & \textbf{0.0013} & \textbf{0.0096} & \textbf{0.0549} & \textbf{3.77e-5} & \textbf{1.15e-4} \\
         \hline
         \multirow{4}{*}{\shortstack{\textbf{Real} \\ \textbf{Data}}} & DPM (GT) & 0.0371 & 1.0390 & 0.1360 & 0.3123 & 0.0534 & 0.1795 & 0.2428 & 0.5644 \\
         & DPM (+20\%) & 0.0339 & 1.0346 & 0.1306 & 0.3178 & 0.0322 & 0.0889 & 0.2400 & 0.5575 \\
         & DPM (-20\%) & 0.0347 & 1.0259 & 0.1342 & 0.3323 & 0.0471 & 0.1397 & 0.2384 & 0.5543 \\
         & DDM (ours) & \textbf{0.0312} & \textbf{0.7976} & \textbf{0.0730} & \textbf{0.1837} & \textbf{0.0187} & \textbf{0.0788} & \textbf{0.1827} & \textbf{0.3840} \\
         \hline
    \end{tabular}}
    \caption{Open-Loop Model Performance}
    \label{tab:sim_rmse}
\end{table}
Table \ref{tab:sim_rmse} shows the RMSE, $\epsilon_{max}$, ADE, and FDE for the DDM and DPM variants on $\mathcal{D}^{\text{test}}_{\text{sim}}$ and $\mathcal{D}^{\text{test}}_{\text{real}}$. 
In simulation, the DDM outperforms all variants of the DPM by over 3 orders of magnitude in terms of both RMSE and $\epsilon_{max}$ on $v_x$ and $\omega$, and 2 orders of magnitude for RMSE and $\epsilon_{max}$ on $v_y$. 
On real data, DDM surpasses the DPM and its variants in terms of RMSE and $\epsilon_{max}$ by over $8\%$ for $v_x$, $56\%$ for $v_y$, and $53\%$ for $\omega$ in comparison to the best DPM.
DDM's ability to predict velocities with high accuracy translates into more precise future position predictions, as evidenced by its lowest ADE, and FDE across both simulated and real data.
Even when given the ground-truth value of $I_z$, the DPM still exhibited ADE and FDE nearly 2 orders of magnitude worse for both metrics.
The best DPM on the real data performed worse as well, with 26\% higher ADE and 36\% higher FDE.


\begin{figure}
\vspace{4pt}
    \centering
    \includegraphics[width=\columnwidth]{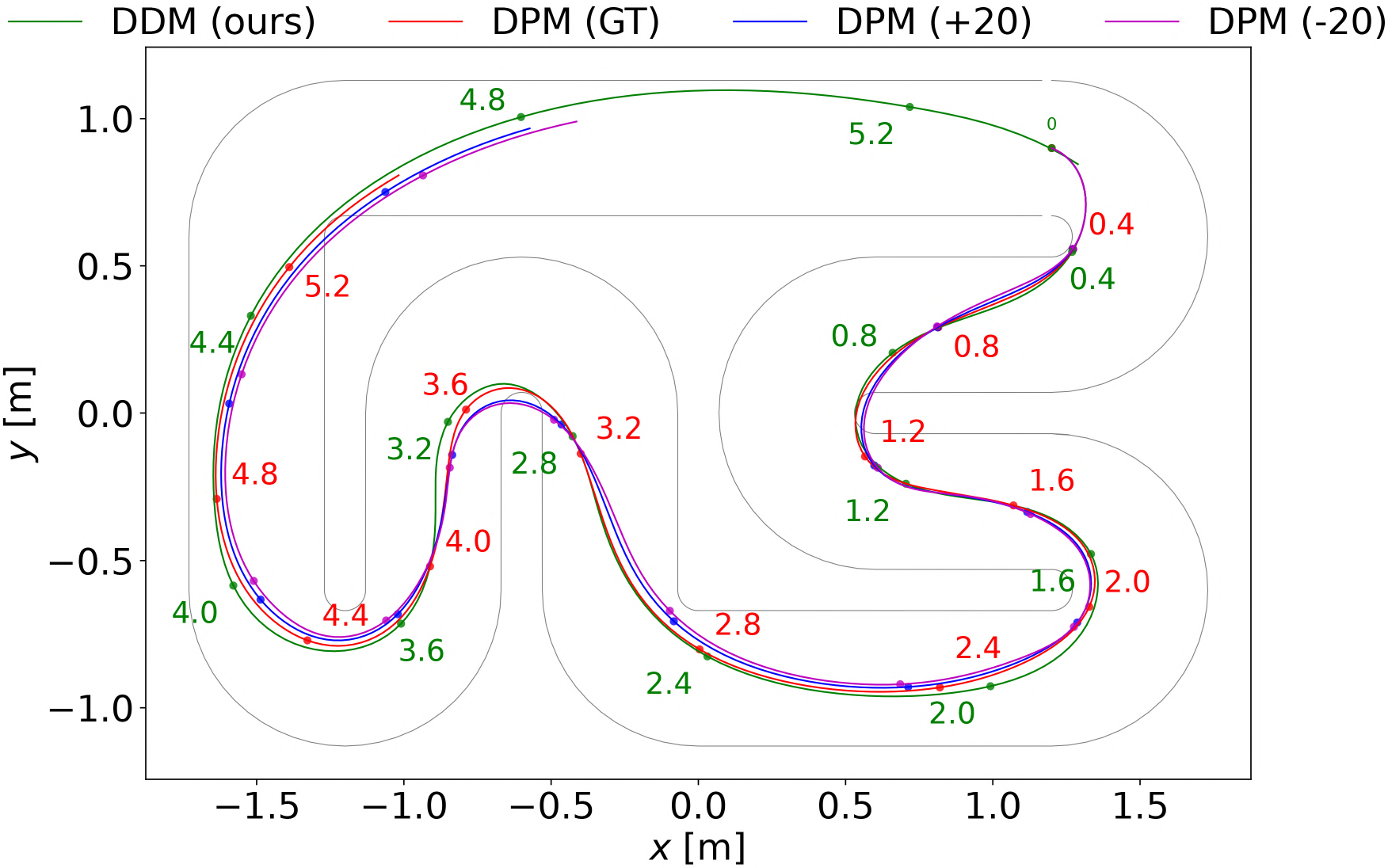}
    \caption{Timestamped trajectories for MPC run using estimated coefficients from DPM (GT, +20, -20) and DDM in the vehicle dynamics simulator.}
    \label{fig:mpc-comparison}
\end{figure}
The models were also compared (shown in Table \ref{tab:coeffs}) by examining their average Pacejka coefficient estimates to evaluate generalization and alignment with physics. Known simulation coefficients provided a benchmark for the simulation data, while nominal coefficient ranges informed the assessment for the real data.
The DPM's estimated Pacejka coefficients for both simulation and real data diverged significantly from anticipated values, being either an order of magnitude different or contrary in sign.
Conversely, the DDM produced estimates closely matching ground-truth values (highlighted in Table~\ref{tab:coeffs}) in simulation including accurately gauging the vehicle’s moment of inertia $I_z$.
Furthermore, the Physics Guard layer ensured that the coefficients derived from $\mathcal{D}^{\text{train}}_{\text{real}}$ generalize to differing track conditions in $\mathcal{D}^{\text{test}}_{\text{real}}$.

\begin{table}[]
    \centering
    \setlength{\tabcolsep}{3pt}
    \resizebox{\textwidth}{!}{
    \begin{tabular}{c l c c c c c c c c c}
        \hline
        && \boldmath{$B_f$} & \boldmath{$C_f$} & \boldmath{$D_f$} & \boldmath{$E_f$} & \boldmath{$B_r$} & \boldmath{$C_r$} & \boldmath{$D_r$} & \boldmath{$E_r$} & \boldmath{$I_z$} \\
        \hline
        \multirow{5}{*}{\shortstack{\textbf{Sim} \\ \textbf{Data}}} & Actual & 5.579 & 1.200 & 0.192 & -0.083 & 5.385 & 1.269 & 0.173 & -0.019 & 2.78e-5 \\
         & DPM (GT) & -4.010 & -0.616 & 0.499 & 1.554 & -1.897 & -0.790 & 0.763 & 8.542 & NA \\
         & DPM (+20\%) & -1.455 & 0.654 & -0.646 & 2.335 & 1.283 & -0.792 & -0.763 & 5.920 & NA \\
         & DPM (-20\%) & -1.156 & 0.614 & -0.662 & 5.329 & -1.175 & 0.695 & -0.725 & 9.823 & NA \\
         & DDM (ours) & \textbf{5.566} & \textbf{1.203} & \textbf{0.192} & \textbf{-0.081} & \textbf{5.505} & \textbf{1.237} & \textbf{0.174} & \textbf{-0.070} & \textbf{2.78e-5} \\
         \hline
         \multirow{5}{*}{\shortstack{\textbf{Real} \\ \textbf{Data}}} & Range & 5-30 & 0.5-2 & 100-10000 & -2-0 & 5-30 & 0.5-2 & 100-10000 & -2-0 & 500-2000 \\
         & DPM (GT) & \cellcolor{green!25} 12.845 & 12.356 & \cellcolor{green!25} 2033.892 & 23.873 & \cellcolor{green!25} 8.377 & 7.892 & -1327.340 & 4.291 & NA \\
         & DPM (+20\%) & \cellcolor{green!25} 5.763 & -28.338 & -1249.052 & 27.035 & 0.095 & 9.587 & \cellcolor{green!25} 863.780 & 12.189 & NA \\
         & DPM (-20\%) & \cellcolor{green!25} 11.739 & 12.335 & \cellcolor{green!25} 1109.978 & 49.836 &  4.823 & 3.056 & -703.947 & 28.165 & NA \\
         & DDM (ours) & \cellcolor{green!25} 6.630 & \cellcolor{green!25} 1.047 & \cellcolor{green!25} 3451.427 & \cellcolor{green!25} -1.051 & \cellcolor{green!25} 15.426 & \cellcolor{green!25} 0.777 & \cellcolor{green!25} 5335.260 & \cellcolor{green!25} -0.733 & \cellcolor{green!25} 1938.432 \\
         \hline
    \end{tabular}}
    \caption{Comparison of Estimated Model Coefficients}
    \label{tab:coeffs}
\end{table}

\subsection{Closed-Loop Testing}

\begin{table}
    \centering
    \vspace{6pt}
    \setlength{\tabcolsep}{3pt}
    \resizebox{\textwidth}{!}{
    \begin{tabular}{c c c c}
        \hline
        \textbf{Model} & \textbf{Lap Time (s)} & \textbf{Average Speed (m/s)} & \textbf{Track Violations} \\
        \hline
        DPM (GT) &  6.16 & 1.721 & 2 \\
        DPM (+20) & 6.00 & 1.732 & 3 \\
        DPM (-20) & 5.94  & 1.728 & 3 \\
        DDM (ours) & \textbf{5.38} & \textbf{2.010} & \textbf{0} \\
        \hline
    \end{tabular}}
    \caption{Closed-Loop Racing Analysis}
    \label{tab:racing_analysis}
\end{table}

Closed-loop testing and comparison with DPM was performed using the MPC implementation described in Section~\ref{ssec:mpc}.
For DPM, a PID controller was used for throttle control (as opposed to a sampling-based MPC controller in the DPM paper).
Figure \ref{fig:mpc-comparison} plots the time annotated traces for each model. 
Racing-relevant metrics such as lap time, average speed, and track boundary violations are reported in Table~\ref{tab:racing_analysis} for a single lap with an initial $v_x$ of $0.1 \si{\meter\per\second}$.
DDM results in the fastest lap time of $5.38$s while incurring $0$ boundary violations, and achieving the highest average speed of over $2\si{\meter\per\second}$ (naively extrapolated to 311 km/h as this is a 1:43 scale simulation).
The inference time for DDM measured on a GeForce RTX 2080 Ti was approximately $700\si{\hertz}$, making it suitable for real-time implementation. 

\subsection{Hyperparameter Tuning}

To ensure a fair comparison, rigorous hyperparameter tuning was conducted for both the DDM and DPM.
Table {\ref{tab:tuned_hyperparams}} presents the optimal hyperparameters for each model, derived from each dataset, and includes the search ranges utilized.
The learning rate, batch size, number of hidden layers, size of hidden layers, and the number of historical timesteps, $\tau$, were all considered hyperparameters. 
Exploration of network architectures included fully connected networks and Gated Recurrent Units (GRU), with GRU models outperforming.
Both models used the Mish activation function for the hidden layers and were optimized using Adam.

\begin{table}[]
    \centering
    \resizebox{\textwidth}{!}{
    \begin{tabular}{c c c c c c c c}
        \hline
         && \textbf{GRU Layers} & \textbf{History ($\tau$)} & \textbf{Hidden Layers} & \textbf{Layer Size} & \textbf{Batch Size} & \textbf{Learning Rate}  \\
         \hline
         \multicolumn{2}{c}{\textbf{Ranges}} & 0 - 16 & 0 - 16 & 0 - 16 & 16 - 512 & 2 - 64 & 1e-4 - 1e-2 \\
         \hline
         \multirow{2}{*}{\shortstack{\textbf{Sim} \\ \textbf{Data}}} & DPM & 8 & 10 & 2 & 108 & 16 & 2.8e-3 \\
         & DDM & 7 & 5 & 16 & 436 & 2 & 1.4e-4 \\
         \hline
         \multirow{2}{*}{\shortstack{\textbf{Real} \\ \textbf{Data}}} & DPM & 5 & 16 & 1 & 254 & 16 & 4.4e-4 \\
         & DDM & 3 & 15 & 2 & 188 & 64 & 1.9e-3 \\
         \hline
    \end{tabular}}
    \caption{Tuned Hyperparameters for Simulation and Real Data}
    \label{tab:tuned_hyperparams}
\end{table}

\section{Conclusion}

We present Deep Dynamics, the first PCNN vehicle dynamics model designed for use in autonomous racing. The network is capable of learning coefficients for a dynamic single-track vehicle model that accurately predicts the vehicle's movement with little knowledge of its physical properties. The Physics Guard layer ensures that these coefficients remain in their physically meaningful range to guarantee the network obeys the laws of physics and improves generalization to unseen data. Our proposed work outperforms its predecessor in both open and closed-loop performance in a vehicle dynamics simulator and data collected from a real-world, full-scale autonomous racecar. 
Future work includes further closed-loop testing, starting with a high-fidelity racing simulator, leading to deploying Deep Dynamics on a full-scale autonomous Indy racecar to demonstrate real-time performance and feasibility.
In addition, in this work, the single-track model coefficients in the simulator were held constant, but we are also interested in investigating the performance of the DDM when these coefficients vary with time.

\bibliographystyle{unsrt}
\bibliography{references,mb}

\end{document}